\documentclass{article}

\usepackage{microtype}
\usepackage{graphicx}
\usepackage{subfigure}
\usepackage{booktabs} 
\usepackage{amsmath}

\usepackage{hyperref}

\usepackage[accepted]{icml2021}

\icmltitlerunning{Scientific Reasoning: Assessment of Multimodal Generative LLMs}
\begin{document}

\twocolumn[
\icmltitle{Scientific Reasoning: \\
           Assessment of Multimodal Generative LLMs}

\icmlsetsymbol{equal}{*}

\begin{icmlauthorlist}
\icmlauthor{Florian Dreyer}{equal,us}
\icmlauthor{Ekaterina Kolos}{equal,us}
\icmlauthor{Daria Matiash}{equal,us}
\end{icmlauthorlist}

\icmlaffiliation{us}{University of Stuttgart}

\icmlcorrespondingauthor{Florian Dreyer}{florian.dreyer03@icloud.com}
\icmlcorrespondingauthor{Ekaterina Kolos}{ekaterina.kolos@ims.uni-stuttgart.de}
\icmlcorrespondingauthor{Daria Matiash}{daria.matiash@ims.uni-stuttgart.de}

\icmlkeywords{Scientific Reasoning, Prefix Tuning, Knowledge Distillation}

\vskip 0.3in
]

\printAffiliationsAndNotice{\icmlEqualContribution} 

\begin{abstract}
Large language models (LLMs) can answer questions and reason about complex tasks, also from the scientific domain. We assess several multimodal LLMs (MLLMs) on ScienceQA and find that Gemini models show the highest accuracy with little context, and the highest textual similarity to human explanations with richer context. Adapter-tuning of smaller MLLMs did not lead to any reliable performance. Training from Gemini outputs consistently underperformed training from the original data. 
\end{abstract}

\section{Introduction}
\label{introduction}

Scientific reasoning involves AI models answering questions by interpreting scientific texts, diagrams, and data.

Recent developments in foundation models, such as OpenAI \cite{OpenAI2024o1} o1/o3 or R1 from DeepSeek \cite{deepseek-r1}, have shown remarkable performance in challenges that require contextual awareness and reasoning.
However, these models are often resource-consuming, which limits their scalability and accessibility for broader applications. This leads us to our research questions:

\paragraph{RQ1}
\label{def:rq1}
How can large multimodal LLMs deal with multimodal scientific reasoning?
%We in fact answer three questions: 
% 1) how well they perform in scientific question answering (QTCH), 
% 2) whether they are capable of reasoning (QTCHL) 
% 3) how well they use in-context information (=solution) to produce expected useful generation

\paragraph{RQ2}
\label{def:rq2}
How do Prefix Tuning and LoRA affect the reasoning capabilities of smaller pre-trained models?

\paragraph{RQ3}
\label{def:rq3}
How good is knowledge distillation with adapter methods compared to training on manually annotated data?

We first want to evaluate how six large front-tier LLMs perform on the \textsc{ScienceQA} dataset \cite{lu2022learn} using several metrics. Following that, we evaluate how well Prefix Tuning \cite{li2021prefix} and LoRA \cite{hu2021lora} perform for fine-tuning two smaller LLMs on this dataset. Finally, we perform knowledge distillation using Prefix Tuning and LoRA and compare how the distilled models compare to the previously fine-tuned models.

\section{Background and Related Work}
\label{related-work}

\begin{figure} [H]
    \centering
    \includegraphics[width=1\linewidth]{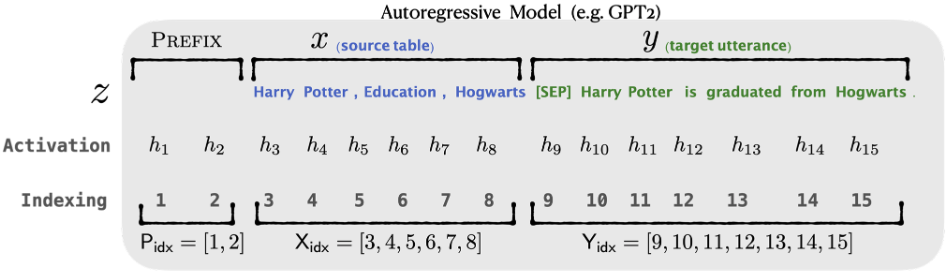}
    \caption{Intuition behind Prefix Tuning. Source: \cite{li2021prefix}}
    \label{fig:prefix_tuning}
\end{figure}

\paragraph{Prefix Tuning}
Prefix Tuning, as introduced in \cite{li2021prefix}, is a parameter-efficient fine-tuning method that freezes the pre-trained model's parameters and trains a small, learnable "prefix" as illustrated in figure \ref{fig:prefix_tuning} that guides the model during task-specific inference. This allows the model to adapt to new tasks while preserving its general pre-trained knowledge.
By significantly reducing the number of trainable parameters, Prefix Tuning is especially advantageous in resource-constrained scenarios or when deploying models for multiple tasks. Studies have shown that it achieves performance comparable to full fine-tuning in many applications, such as natural language generation and classification.

\paragraph{LoRA}
Low-Rank Adaptation (LoRA) \cite{hu2021lora} is a parameter-efficient fine-tuning technique designed to adapt pre-trained models to downstream tasks with minimal computational overhead. LoRA introduces trainable low-rank matrices into the model's attention layers while keeping the original weights frozen. These matrices encode task-specific knowledge, allowing the model to be fine-tuned with a significantly smaller number of parameters compared to traditional fine-tuning. LoRA is particularly effective in scenarios where memory efficiency is critical as it avoids modifying or storing the full set of model weights. 
Studies have shown that LoRA achieves competitive performance on various NLP tasks, such as machine translation and text classification \cite{mao2025survey}.

\paragraph{Knowledge distillation (KD)} allows to obtain smaller models capable of successfully following the behavior of larger teacher models. This is particularly useful for privacy reasons (mobile AI apps) and when access to large models in the cloud is not guaranteed (e.g., in cars). The teacher model can be used to intelligently select examples on which the student model is trained (dataset distillation) \cite{yu2023dataset}, or provide negative samples to show the student what incorrect answers or reasoning paths should be avoided to improve accuracy \cite{li2024turning}.
%(c.f. contrastive CoT). 
Training on a chain-of-thought reasoning path of a larger model was also shown to be a way to obtain a small student model that replicates the reasoning capabilities of the teacher on downstream tasks \cite{magister2022teaching}. 
% which is close to \textit{response-based KD} where the student model mimics the output of the teacher. 
We follow a similar approach and perform \textit{response-based knowledge distillation} \cite{gou2021knowledge}, learning to mimic the output of the teacher.
%Alternatively, with \textit{feature-based KD} the knowledge from specific layers of the teacher model is distilled into the student model \cite{sepahvand2022teacher}, while the student model's structure may be a quantized, simplified, or condensed version of the teacher's architecture \cite{gou2021knowledge}. 

\paragraph{Models} We experiment with the following foundation models: 
\texttt{Pixtral-12b-2409} \cite{agrawal2024pixtral}, \texttt{LLaVA-1.5-7b-hf} \cite{liu2024improved}, \texttt{Gemini 1.5 Flash}, \texttt{Gemini 1.5 Flash 8B} \cite{team2024gemini}, \texttt{GPT 4}, \texttt{GPT 4o mini} \cite{achiam2023gpt}, %and two small MLLMs: 
\texttt{Qwen2-VL-2B-Instruct} \cite{wang2024qwen2}, \texttt{paligemma-3b-pt-224} \cite{beyer2024paligemma}.

\section{Dataset}
\label{dataset}
This study is based on the \textsc{ScienceQA} dataset \cite{lu2022learn}. \textsc{ScienceQA} is a benchmark for multimodal reasoning in science, consisting of more than 21,000 questions across topics such as natural science or social science. 
Each question includes textual prompts, optional visual aids (e.g., diagrams, charts), answer options, detailed explanations, and lecture context. The dataset spans various difficulty levels, enabling evaluation of models on both basic and advanced scientific reasoning, ranging from elementary to high-school level questions.

The dataset contains some notable irregularities. For some datapoints, the image data is missing (sometimes when it is required to solve the task, sometimes when the answer can be deduced from text alone).
Apart from the image field, missing values can occur in lecture and solution fields (about 9\% and 15\% respectively). As about 50\% of the datapoints are missing an image, we decide to process both text-only and text+image datapoints similarly with multimodal models, attaching the image to the prompt if it was present in the data. When the image was not available, we generated a blank empty image as a placeholder. 

\section{Methodology}
\label{section:methodology}
In the following we present the methodology we used to examine the three research questions (RQ).
\paragraph{RQ1} We benchmark six front-tier LLMs using accuracy and the average of five text similarity metrics introduced in the metrics section \ref{section:metrics}. 
To investigate how additional information and the correct solution in the prompt influence the performance, we use four different prompt settings: 
\begin{enumerate}
	\item \textbf{question - choices - hint - image - task}
	\item \textbf{question - choices - hint - image - task + lecture}
	\item \textbf{question - choices - hint - image - task + lecture + solution}
	\item \textbf{question - choices - hint - image - task + solution}
\end{enumerate}
Each model is benchmarked on all four settings using the \textsc{ScienceQA} validation data. We use the validation split instead of the test split, since we later use the results to choose a champion teacher model. This decision cannot be made using the test data since it is utilized to evaluate knowledge distillation performance.

\paragraph{RQ2} For the comparison of Prefix Tuning and LoRA we fine-tune two smaller multimodal LLMs
using the two techniques separately. As the label to train on, we choose the solution to enable the adapters to learn more of the reasoning. We then compare the performance of the four fine-tuned models with the zero-shot performance of the two base models on \textsc{ScienceQA} test data.

\paragraph{RQ3} 
We further investigate the impact of knowledge distillation on the performance of adapter tuning. Would learning from the outputs of a champion teacher LLM drop the performance significantly and consistently, compared to learning from the original data? For this, we obtain outputs of the champion teacher model on the train partition of \textsc{ScienceQA}. We train 2 different adapters on 2 different students on this data and on the original train partition of \textsc{ScienceQA}, and compare the text similarity results in section \ref{section:experiments:KD}. 

We compare the performance of the eight resulting models on the \textsc{ScienceQA} test data. 

\section{Metrics}
\label{section:metrics}

We evaluate the models' performance with QA accuracy.  
The reasoning steps were evaluated with semantic similarity metrics, adopted from machine translation, such as BLEU, METEOR, ROUGE, and cosine similarity.
\subsection{Multiple choice Evaluation}
Owing to the simplicity of the test format, only the accuracy score is computed, following the original evaluation strategy by \cite{lu2022learn}. To extract the answers from the output, we prompt the models to output in JSON format with "answer" and "solution" as keys.
\subsection{Answer Reasoning Evaluation}
Due to the peculiarity of scientific texts and approaches to the evaluation of automatically generated texts, the following evaluation approaches were chosen.

\paragraph{BLEU}
BLEU-score \cite{papineni2002bleu} can measure how closely the model's generated solutions aligns with the human-authored explanation, calculating modified \textit{n}-gram precisions adjusted by brevity penalty. BLEU-1 measures if the model uses the right scientific terms or key words (e.g., \textit{"photosynthesis"},  \textit{"temperature"}). However, it ignores word order, 
so it can’t evaluate reasoning or explanation quality. That is why BLEU-4 score is also computed to capture both vocabulary usage and phrase structures.
By computing both BLEU-1 and BLEU-4, we balance term accuracy with linguistic structure, ensuring a more reliable evaluation of the model’s explanatory capabilities in scientific reasoning.

\paragraph{ROUGE}
ROUGE is a set of metrics introduced in \cite{lin2004rouge}, again, to estimate the quality of a generated hypothesis compared to one or more golden references. 
ROUGE-1 and ROUGE-2, like BLEU-\textit{n}, count overlaps of \textit{n}-grams between the hypothesis and the reference, calculating not only precision, but also recall and F1-score.
Following the evaluation strategy in \cite{lu2022learn}, we adopt ROUGE-L score \cite{lin2004rouge}, which calculates the longest common subsequence between the hypothesis and the reference. 
As opposed to ROUGE-1 and ROUGE-2, this score would assign a higher value to e.g. \textit{"the theory which Einstein proposed"} than to \textit{"Einstein proposed the theory" } for a reference like \textit{"the theory that Einstein proposed"}.
This score, however, would not capture synonyms by default, which is why we do not rely on this score alone. 
\footnote{We performed our evaluation following the scripts proposed by the authors of \textsc{ScienceQA} for consistent comparison. Upon more careful investigation done at the stage of writing this report, we found out that the package used for ROUGE was https://pypi.org/project/rouge/, which does not calculate the score consistently with the algorithm proposed by \cite{lin2004rouge}. A better implementation, that is more accurate to it, would be this one: https://pypi.org/project/rouge-score/.}

\paragraph{METEOR}
METEOR \cite{banerjee2005meteor} provides a more fine-grained and linguistically informed approach to evaluating text similarity, balancing precision and recall, and encouraging correct alignment of semantically meaningful components through fragmentation penalty.  
Later versions of METEOR introduce stemming, synonym matching, as well as fine-grained weights for individual languages for even better agreement with human judgments.

\paragraph{Cosine similarity with Sentence Transformers}
\texttt{all-MiniLM-L6-v2} was utilized in Sentence Transformers library in our evaluation strategy.
Unlike \texttt{SciBERT}, which is specialized for scientific text and could serve as a sentence embedder, \texttt{all-MiniLM-L6-v2} was pre-trained on diverse datasets, allowing it to better handle variations in phrasing and style found in the Science QA Dataset.
Its broader linguistic adaptability ensures that the evaluation considers both semantic accuracy and logical flow, which is crucial for assessing reasoning across different scientific domains.

\paragraph{Overall score}
To make the performance of the models easier for comparison, we computed an overall score of the textual similarity metrics. In order to obtain a numeric description of the model's performance for these text similarity metrics, we computed the average of them (having normalized cosine similarity to the range [0;1]). 

\section{Experiments}
\label{experiments}
In the following, we present the experiments to answer the three RQs.

\subsection{Benchmarking large LLMs}
As described in \ref{section:methodology} we benchmarked six large LLMs on the validation part of the dataset. Our experiments resulted in the following results for accuracy:
\begin{figure} [h]
    \centering
    \includegraphics[width=1\linewidth]{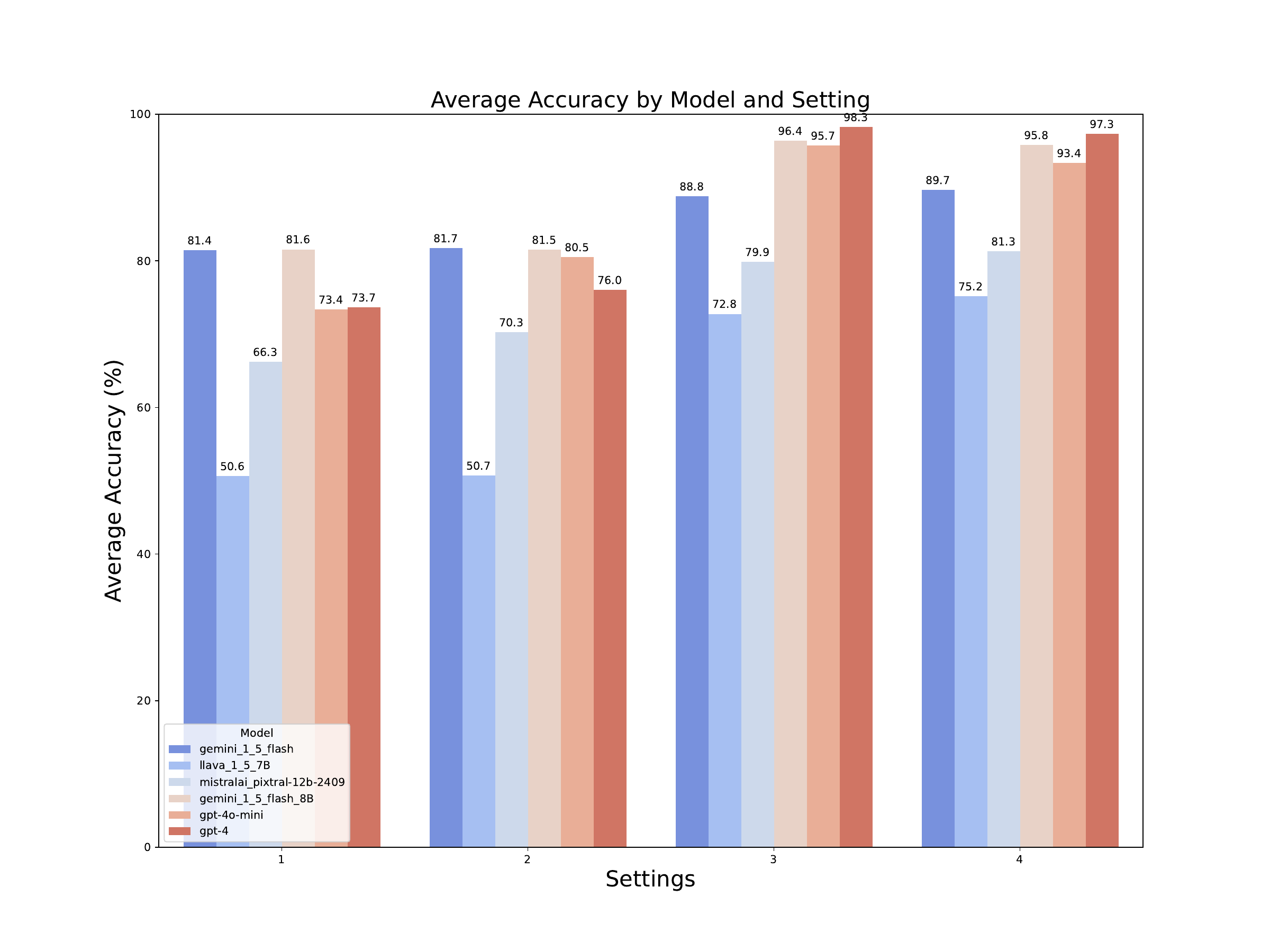}
    \label{fig:benchmark_accuracy}
    \caption{Accuracy scores in answer generation by LLMs. Benchmarking.}
\end{figure}

For the overall score as introduced in the \ref{section:metrics} the experiments resulted in the following scores:
\begin{figure} [H]
    \centering
    \includegraphics[width=1\linewidth]{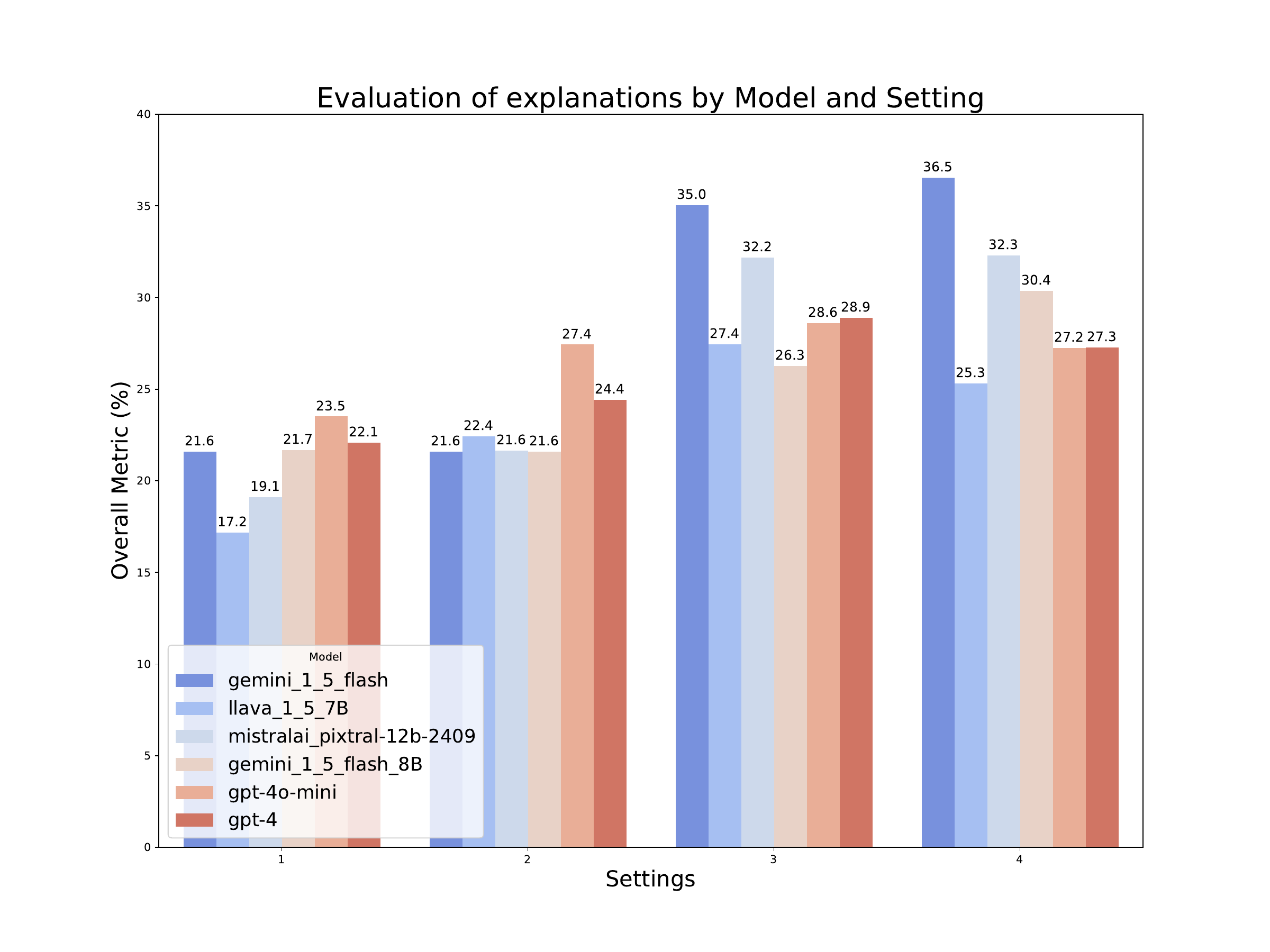}
    \label{fig:benchmark_overall}
    \caption{Overall scores in reasoning by LLMs. Benchmarking.}
\end{figure}
For the individual textual similarity metrics, we got the following results:
\begin{table} [H]
\vskip 0.15in
\begin{center}
\begin{small}

\begin{tabular}{llcccccc}
\toprule
model & s. & bl-1 & bl-4 & r. & m. & cos. & overall \\
\midrule
\textsc{Gemini} & 1 & 0.04 & 0.01 & 0.28 & 0.14 & 0.80 & 21.59 \\
 & 2 & 0.04 & 0.01 & 0.28 & 0.14 & 0.80 & 21.58 \\
 & 3 & 0.12 & \textbf{0.12} & \textbf{0.52} & \textbf{0.30} & \textbf{0.84} & \textbf{35.03} \\
 & 4 & 0.14 & \textbf{0.14} & \textbf{0.53} & \textbf{0.32} & \textbf{0.85} & \textbf{36.55} \\
\midrule
\textsc{LLaVA} & 1 & 0.03 & 0.01 & 0.24 & 0.09 & 0.74 & 17.16 \\
 & 2 & 0.08 & \textbf{0.05} & \textbf{0.33} & 0.06 & 0.80 & 22.43 \\
 & 3 & 0.06 & 0.06 & 0.41 & 0.20 & 0.82 & 27.43 \\
 & 4 & 0.05 & 0.05 & 0.41 & 0.13 & 0.81 & 25.32 \\
\midrule
\textsc{Pixtral} & 1 & 0.05 & 0.01 & 0.23 & 0.11 & 0.78 & 19.11 \\
 & 2 & 0.06 & 0.02 & 0.27 & 0.14 & 0.79 & 21.64 \\
 & 3 & 0.12 & 0.11 & 0.46 & 0.27 & 0.83 & 32.17 \\
 & 4 & 0.11 & 0.11 & 0.47 & 0.27 & 0.83 & 32.29 \\
\midrule
\textsc{Gemini}  & 1 & 0.05 & 0.01 & 0.28 & 0.14 & \textbf{0.81} & 21.68 \\
\textsc{8B} & 2 & 0.05 & 0.01 & 0.27 & 0.14 & 0.81 & 21.59 \\
 & 3 & 0.06 & 0.04 & 0.37 & 0.19 & 0.82 & 26.26 \\
 & 4 & 0.08 & 0.08 & 0.44 & 0.24 & 0.84 & 30.37 \\
\midrule
\textsc{GPT-4o} & 1 & \textbf{0.14} & \textbf{0.03} & 0.28 & \textbf{0.17} & 0.78 & \textbf{23.51} \\
\textsc{-mini} & 2 & \textbf{0.17} & 0.04 & 0.31 & \textbf{0.21} & \textbf{0.82} & \textbf{27.43} \\
 & 3 & \textbf{0.18} & 0.05 & 0.33 & 0.22 & 0.82 & 28.61 \\
 & 4 & \textbf{0.17} & 0.04 & 0.31 & 0.21 & 0.82 & 27.24 \\
\midrule
\textsc{GPT-4} & 1 & 0.11 & 0.02 & \textbf{0.30} & 0.16 & 0.76 & 22.09 \\
 & 2 & 0.13 & 0.04 & 0.32 & 0.18 & 0.78 & 24.41 \\
 & 3 & 0.16 & 0.06 & 0.35 & 0.23 & 0.82 & 28.88 \\
 & 4 & 0.14 & 0.04 & 0.33 & 0.21 & 0.82 & 27.29 \\
\bottomrule
\end{tabular}

\end{small}
\end{center}
\caption{Metrics on scientific reasoning by LLMs. \textit{s.} - setting, \textit{bl-1} - BLEU-1, \textit{bl-4} - BLEU-4, \textit{r.} - ROUGE-L, \textit{m.} - METEOR, \textit{cos.} - cosine similarity, \textit{overall} - overall metric in \%.}
\end{table}

\paragraph{Accuracy}

%1 QTCH: (81.4+50.6+66.3+81.6+73.4+73.7)/6 = 71.167
%2 QTCHL: (81.7+50.7+70.3+81.5+80.5+76)/6 = 73.45
%3 QTCHLS: (88.8+72.8+79.9+96.4+95.7+98.3)/6 = 88.65
%4 QTCHS: (89.7+75.2+81.3+95.8+93.4+97.3)/6 = 88.783

The highest performance was observed in settings that included solutions, indicating that the models were generally capable of extracting relevant information effectively. 
We noticed that setting 3 including the lecture did yield worse results for most models compared to the lecture-free setting 4.
\texttt{GPT-4} and \texttt{Gemini-1.5-flash-8B} consistently received highest ratings in scenarios in which lecture and/or solution information was included. The \texttt{Gemini} family of models and  \texttt{Pixtral-12b-2409} model showed better robustness in many scenarios and came in second overall. Notably, \texttt{Gemini} models achieved an average accuracy advantage of 8\% over \texttt{GPT-family} models in the "pure" task context (without lecture or solution information). The \texttt{LLaVA-1.5-7b-hf}, on the other hand, demonstrated the lowest average accuracy scores, particularly in environments without lecture and solution data. For setting 1, which was considered as the most important for further experiments, \texttt{Gemini} models performed with highest accuracy scores.

\paragraph{BLEU-1}
Across settings, the highest BLEU-1 scores were mostly observed in settings 3 and 4 (those incorporating solution or lecture+solution information), indicating that models performed better with richer contextual data. The \texttt{GPT-4o-mini} model demonstrated superior performance, surpassing \texttt{GPT-4}, in achieving the highest results. Lower scores in settings 1 and 2 suggest that models, particularly \texttt{LLaVA-1.5-7b-hf}, may have had trouble correctly aligning their outputs with reference words when given less auxiliary information. 

\paragraph{BLEU-4}
\texttt{GPT-4} models had higher scores in comparison to other models, which represents a better capability of information extraction. \texttt{Gemini}, at the same time, gained the highest score in the setting without solutions, which indicates the ability of extracting knowledge from massive textual data.

\paragraph{METEOR}
METEOR scores reveal how well the models capture fluency, grammar, and word-level alignment with the references. Similar to cosine similarity, \texttt{GPT-4o-mini} leads with the highest METEOR scores. Notably, \texttt{Pixtral-12b-2409} achieves relatively high METEOR scores in settings 3 and 4, highlighting its ability to produce fluent outputs. When concrete tasks without any additional helpful material are given to the models as an input, \texttt{GPT-4o-mini} model shows a capability of generating concise, fluent answers with correct relevant terminology and explanations.

\paragraph{ROUGE}
The highest ROUGE scores were observed in setting 3, where both lectures and solutions were available, highlighting the importance of comprehensive input for producing informative responses. Setting 1 showed lower ROUGE scores across almost all models, reflecting limited informativeness when models were provided with minimal context.  \texttt{GPT} family of models had highest ROUGE-scores on settings without solutions. Outputs of \texttt{GPT-4} without knowledge of the correct answer had a higher ROUGE score than \texttt{GPT-4o-mini} given solutions in the input.

\paragraph{Cosine similarity}
Cosine similarity highlights the semantic alignment between the model outputs and the correct answers. {Gemini-1.5-flash} consistently achieved the highest similarity scores, particularly in settings 3 and 4, indicating strong alignment with the reference answers. \texttt{GPT-4} variants maintain steady scores across all settings, demonstrating robustness, while \texttt{Pixtral-12b-2409} and \texttt{LLaVA-1.5-7b-hf} show more variability, suggesting sensitivity to specific settings. Overall, cosine similarity shows that {Gemini-1.5-flash} excels in semantic understanding across the dataset.

\paragraph{Accuracy and Overall score}
The overall results indicate that the \texttt{GPT} and \texttt{Gemini} model families demonstrate exceptional ability to extract accurate answers. Notably, the \texttt{Gemini} models exhibited superior performance on datasets lacking relevant lecture information in most of the metrics. The second-best performance on "pure" (without solutions and/or lectures) datasets was achieved by \texttt{GPT-4o-mini}, with only a slight difference in overall score.

\paragraph{Summary}
Overall, the findings demonstrate how important enriched contextual data, such as lectures and solutions, in enhancing model performance across all metrics. The \texttt{GPT-4o-mini} model outperformed \texttt{GPT-4} in multiple instances, indicating the efficacy of smaller, more targeted designs, even if \texttt{GPT-4} continuously received excellent accuracy and scores for reasoning. While \texttt{LLaVA-1.5-7b-hf} model failed without extra input, highlighting its dependence on extensive contextual knowledge to function well, the \texttt{Gemini} family of models demonstrated remarkable robustness and semantic alignment, particularly in enriched situations.
Lecture information doesn’t really affect models’ performance, it can even decrease the ratio of correct answers in some cases.

\subsection{Prefix Tuning and LoRA}
For training the Prefix Tuning adapters we used 20 epochs for Prefix Tuning as derived from \cite{li2021prefix} and limited by computing resources. 
For training the LoRA adapter we used 10 epochs as the loss flattened after more epochs in first experiments.
After training \texttt{Qwen} and texttt{PaliGemma} using both Prefix Tuning and LoRA we benchmarked the resulting models on the \textsc{ScienceQA} test data. We were not able to measure the accuracy because all four models were not able to output valid answers. For the overall score of the text similarity the experiments resulted in the following scores:
\begin{figure} [H]
    \centering
    \includegraphics[width=1\linewidth]{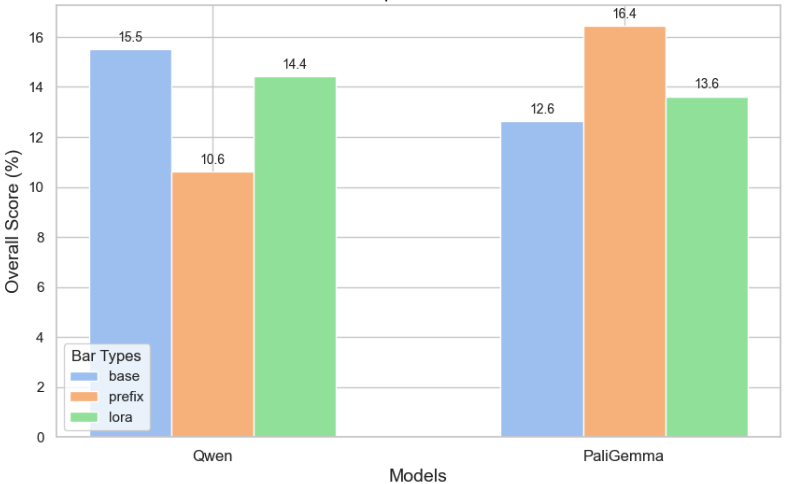}
    \label{fig:prefix_lora_test}
    \caption{Overall score in reasoning by LLMs in base, after Prefix-tuning, and LoRA Adapter-tuning.}
\end{figure}

Overall we cannot see a consistent trend, for \texttt{Qwen}, the base model outperformed both adapters while for \texttt{PaliGemma} Prefix Tuning resulted in the best overall score, while the base model performed the worst. It was noticed that the more epochs of efficient parameter tuning was done, the shorter the text became that was generated by both models. 

\subsection{Knowledge Distillation}
\label{section:experiments:KD}
For training, we use the same number of epochs as for the previous experiments.
For evaluating the perfomance of Knowledge Distillation compared to normal fine-tuning using the adapter methods we trained both \texttt{Qwen} and \texttt{Paligemma} using Prefix Tuning and LoRA separately. All eight trained models were not able to output valid answers, so we will not compare the accuracy. For the overall score of the text similarity, the experiments resulted in the following scores:
\begin{figure}[H]
    \centering
    \includegraphics[width=1\linewidth]{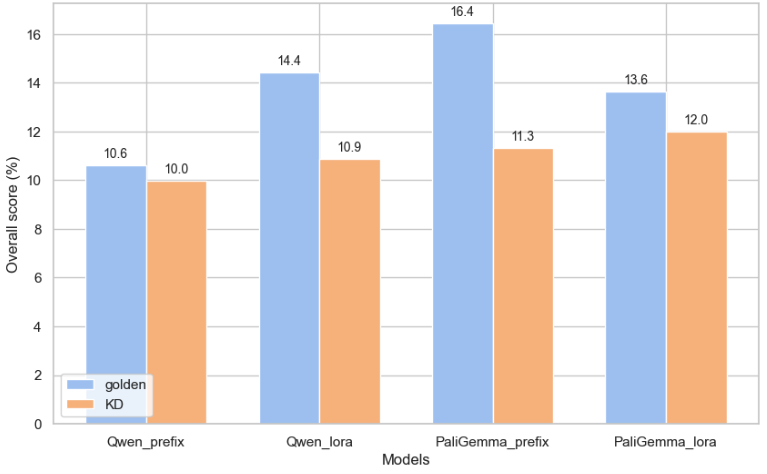}
    \label{fig:KD_overall}
    \caption{Overall score in reasoning by LLMs.}
\end{figure}
%\paragraph{Summary}
Overall we can see a clear outperformance of Knowledge Distillation by the fine-tuned models. We can also notice that Knowledge Distillation performs quite similar for all four models, while the fine-tuning approach has higher variance in performance. We noticed that the LoRA approach outperforms Prefix Tuning for Knowledge Distillation for both models. This is not the case for the fine-tuned models.

\section{Error Analysis}
\label{error-analysis}
\paragraph{Output format}
We noticed that even front-tier SOTA models could sometimes deviate from the expected output format; this error type was much more common in smaller models, despite attempts of controlled decoding. Furthermore, a fine-tuned version of \texttt{Paligemma} would consistently produce outputs with the \textit{answer} only, not giving any \textit{solution}.
For example, for an input asking to output JSON with solution: "\textit{Question: Which of the following organisms is the primary consumer in this food web? $\langle...\rangle$ Choices: ['copepod', 'black crappie', 'bacteria'] $\langle...\rangle$  Instruction:  Please output the answer in JSON style with an answer and a solution field}" the finetuned \texttt{Paligemma} would only output "\textit{The answer is A.}".
Qwen models would sometimes produce Chinese utterances instead of expected English generations. 
%化学变化 - 经过发酵或分解，可以将有机物（如水果）转化为二氧化碳、水和无机盐。
\newline
\newline 
Models would sometimes generate long sequences going beyond a stop token, degenerating into completely irrelevant text or even code. The \texttt{Pixtral} model could finish answering the given task and then generate a new $\langle s\rangle[INST]$ token, after which a new hallucinated task would follow, which the model would later answer.
Finally, standardizing the answer, which, given 2-5 choices, could be a number, a letter, or a string, proved rather difficult, with some 500 answers left unmapped in each experiment. 
\paragraph{Sparse outputs}
Adapter training surprisingly made the outputs completely degenerate into quasi-empty strings, containing only spaces, functional words, numbers, or rare irrelevant or foreign tokens.
We speculate that this results from an overly strict feedback that the models got every time they generated plausible, yet not exactly the same correct explanations. 

\paragraph{Output quality}
Finally, the large LLMs from which the teacher model was selected could not generate perfectly adequate outputs either, being overly vague in explanations or, on the contrary, bringing up irrelevant or hallucinated details. For example, for a task question: "\textit{What's the difference between weather and climate?}" one of the generations was: "\textit{Climate is the pattern of weather in a certain place. It got down to 3°C in Athens, Greece, last night!}", which included irrelevant hallucinated information about the weather in Athens on a specific date. 

\section{Conclusion}
\label{Conclusion}
In conclusion, while big LLMs demonstrate strong capabilities in multimodal scientific question answering and benefit from extracting information from available solutions, their performance in reasoning from lectures often falls short or even declines. On the other hand, small foundation models like \texttt{Qwen2-VL-2B-Instruct} and \texttt{paligemma-3b-pt-224} show limited effectiveness in scientific reasoning tasks, both in zero-shot settings and when fine-tuned with adapter-based methods. Furthermore, inconsistencies in evaluation metrics (zero-shot, Prefix Tuning, LoRA) across these models highlight challenges in establishing reliable performance benchmarks. The poor performance of adapter tuning may be attributed to the current loss function design, suggesting the need for refinement in optimization strategies. LoRA outperforms Prefix Tuning for both student models in the knowledge distillation setting. Lastly, knowledge distillation underperforms when compared to directly training on a curated, high-quality dataset, emphasizing the importance of data quality in achieving robust model performance.
%It was noticed that all four models had more similar performance with knowledge distillation than with fine-tuning. 
\footnote{All our work is available on our GitHub repository: https://github.com/katja-kolos/foundation\_models}

\section{Future work}
\label{Future work}
While we observed that learning from teachers' outputs leads to lower performance compared to learning from human-curated solutions, the precise impact remains to be measured, since we were not able to derive accuracy for the student models. If the teacher's performance is estimated at 80\% of human performance, would models trained on the teacher's outputs achieve 80\% of the performance of those trained on golden data, or could the student's ability to learn from noisy data mitigate this effect?
One could also run experiments with different Knowledge Distillation techniques like the ones introduced in Background (\ref{related-work}) or \cite{gou2021knowledge}.

Another direction for future work could be trying out other adapter architectures, starting from larger student models (which would require more computational resources but probably provide better results), and learning with a less strict loss function.
\newline
We could additionally define a multi-head setup for fine-tuning, with part of the model being responsible for explanation generation and another one for answer prediction.

\section{Acknowledgments}
The authors acknowledge support by the state of Baden-Württemberg through bwHPC.

%\printbibliography
\newpage
\newpage
\bibliography{main}
\bibliographystyle{icml2021}

\end{document}